\documentclass[conference]{IEEEtran}

\usepackage{amsmath,graphicx}
\usepackage{pgfplots}
\pgfplotsset{compat=1.18}
\usepackage{booktabs}
\usepackage{caption}
\usepackage{graphicx}
\usepackage{multirow}
\usetikzlibrary{arrows.meta}
\begin{document}
%
\title{Enhancing AAC Software for Dysarthric Speakers in e-Health Settings: An Evaluation Using TORGO}

\author{
    \IEEEauthorblockN{Macarious Hui\IEEEauthorrefmark{1}, Jinda Zhang\IEEEauthorrefmark{1}, Aanchan Mohan\IEEEauthorrefmark{1}\IEEEauthorrefmark{2}}
    \IEEEauthorblockA{\IEEEauthorrefmark{1}Northeastern University, Vancouver, BC, Canada \\ \{hui.mac, zhang.jinda1,aa.mohan\}@northeastern.edu}
    \IEEEauthorblockA{\IEEEauthorrefmark{2}University of Victoria, Victoria, BC, Canada \\ }
    
}

\maketitle

\begin{abstract}
Individuals with cerebral palsy (CP) and amyotrophic lateral sclerosis (ALS) frequently face challenges with articulation, leading to dysarthria and resulting in atypical speech patterns. In healthcare settings, communication breakdowns reduce the quality of care. While building an augmentative and alternative communication (AAC) tool to enable fluid communication we found that state-of-the-art (SOTA) automatic speech recognition (ASR) technology like Whisper and Wav2vec2.0 marginalizes atypical speakers largely due to the lack of training data. Our work looks to leverage SOTA ASR followed by domain specific error-correction. English dysarthric ASR performance is often evaluated on the TORGO dataset. Prompt-overlap is a well-known issue with this dataset where phrases overlap between training and test speakers. Our work proposes an algorithm to break this prompt-overlap. After reducing prompt-overlap, results with SOTA ASR models produce extremely high word error rates for speakers with mild and severe dysarthria. Furthermore, to improve ASR, our work looks at the impact of n-gram language models and large-language model (LLM) based multi-modal generative error-correction algorithms like Whispering-LLaMA for a second pass ASR. Our work highlights how much more needs to be done to improve ASR for atypical speakers to enable equitable healthcare access both in-person and in e-health settings.\end{abstract}


%
\IEEEpeerreviewmaketitle

\section{Introduction}
\label{sec:intro}
Healthcare professionals rely on augmentative and alternative communication (AAC) software to support telehealth and in-person appointments for patients with cerebral palsy (CP) and amyotrophic lateral sclerosis (ALS)~\cite{handberg2018implementing}. Damage to the nervous system can result in paralysis or weakness of the muscles responsible for speech, leading to dysarthria and atypical speech patterns in individuals with ALS or cerebral palsy. Atypical speakers who are verbal, often prefer to use their own voice to communicate their needs. Modern AAC applications like VoiceItt\footnote{www.voiceitt.com} or our own AAC application SpeakEase~\cite{mohanpowerful} allow for audio input from the speaker with the intention to provide a faithful transcription. Mulfari et al.~\cite{mulfari2023edge} propose a low-power, on-device, deep-learning based isolated word ASR system to work in an ``always-on'' mode for dysarthric speakers with reduced mobility. Such a system has promise to enable communication in healthcare and home settings.

\begin{figure}[htb]
    \centering
    
    \begin{minipage}[b]{1.0\linewidth}
        \centering
        \includegraphics[scale=0.2]{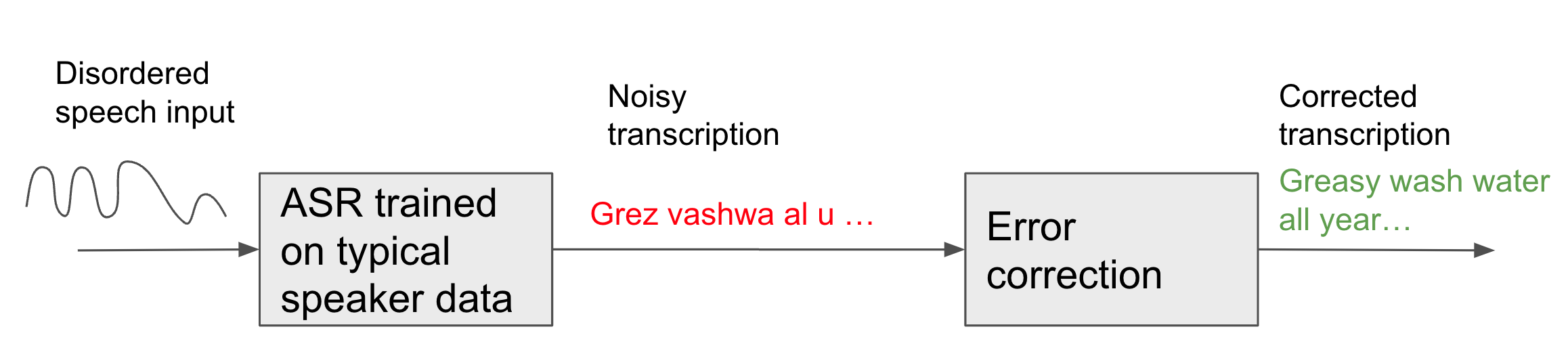}
        \caption{Dysarthric automatic speech recognition followed by error correction}
        \label{fig:dist}
    \end{minipage}
\end{figure}

There is little or no data available on the open domain for atypical speakers. On the other hand web-scale speech datasets like Mozilla Common Voice~\cite{ardila2019common} and GigaSpeech~\cite{chen2021gigaspeech} allow for state-of-the-art speech recognition for typical speakers. Dysarthric speech recognition is a low-resource out-of-domain problem~\cite{li2024crossmodal}. To leverage well-developed typical speaker ASR systems, our work looks for a first pass transcription from such a system followed by error-correction (EC) as show in Figure~\ref{fig:dist}. The figure shows disordered input speech with an imperfect noisy transcription after ASR followed by error-correction. ASR systems are trained with audio and correct transcription pairs. EC systems are trained with inputs consisting of multiple hypotheses of transcribed text (referred to as n-best lists), possibly with errors, with outputs mapping to the correct target text.

\begin{figure} [htb]   
    \begin{minipage}[b]{1.0\linewidth}
        \small
        \centering
        \resizebox{0.8\linewidth}{!}{%
            \begin{tabular}{|l|l|}
                \hline
                1 & \textcolor{blue}{Ref}: he slowly takes a short walk in the open air each day  \\
                  & \textcolor{red}{ASR}: he shlly takes a wall in the week a eh day \\
                  & \textcolor{green}{EC}: he slowly takes a short walk in the open air each day  \\ \hline
                2 & \textcolor{blue}{Ref}: usually minus several buttons  \\
                  & \textcolor{red}{ASR}: usually min sell fold buttons \\
                  & \textcolor{green}{EC}: usually sell fold buttons \\ \hline
                3 & \textcolor{blue}{Ref}: you wished to know all about my grandfather  \\
                  & \textcolor{red}{ASR}: u' wal awarke youar gread fap \\
                  & \textcolor{green}{EC}: you wished to know all about my grandfather \\ \hline
            \end{tabular}%
        }
        \caption{Inference samples for error-correction (\textcolor{green}{EC}) for speaker M05. \textcolor{blue}{Ref} shows the reference transcription, \textcolor{red}{ASR} shows the transcription output which serves as input to the \textcolor{green}{EC} model. Notice how the \textcolor{green}{EC} system has memorized transcripts due to prompt overlap in TORGO.}
        \label{table:inference_samples}
    \end{minipage}

\end{figure}
To evaluate English ASR for dysarthric speakers, a well-known dataset called the TORGO dataset~\cite{rudzicz2012torgo} is widely used. The TORGO dataset for dysarthric speech has data from speakers with either ALS or CP. Other dysarthric ASR databases such as the Nemours corpus\footnote{The authors were unable to obtain a recent copy of this database due to a lack of information on the internet}~\cite{menendez1996nemours}, UASpeech~\cite{hasegawa2006universal} and the HomeService corpus~\cite{nicolao2016framework} are either hard to obtain or largely consist of isolated word utterances. The TORGO dataset is one of the few containing both isloated word and a few sentence level utterances. 

Figure~\ref{table:inference_samples} shows inference examples from our initial experiments of error-correction(\textcolor{green}{EC}) following \textcolor{red}{ASR}. The \textcolor{green}{EC} model memorizes the target transcription without doing any error-correction. This issue stems from the dataset design, which features a significant amount of prompt overlap among the speakers. The research community acknowledges that the TORGO dataset has a very high degree of prompt overlap between speakers~\cite{hermann2020dysarthric, yue2020exploring}. This data leakage prevents the dataset from being used to evaluate ASR and EC algorithms for real-world applications like telehealth and e-health.  

Our work in this paper makes the following contributions: 
\begin{itemize}
    \item Develop an algorithm based on mixed-integer linear programming to partition the TORGO dataset with no prompt overlap with the constraint to minimize data loss. This dataset is called no-prompt overlap TORGO or NP-TORGO. 
    \item Understand the impact of removing prompt overlap on dysarthric ASR performance using SOTA baseline ASR models.
    \item Understand the impact of out-of-domain language modelling using text data from the training utterances from NP-TORGO, and Librispeech~\cite{panayotov2015librispeech} .
    \item Understand the impact of error-correction (EC) without ASR system fine-tuning, with a state-of-the-art cross-modal error-correction system such as Whispering-LLAMA~\cite{radhakrishnan-etal-2023-whispering}.
\end{itemize}

This paper is organized as follows. Section~\ref{prior_work} puts our current work in the context of prior work. Section~\ref{ec_intro} introduces the TORGO dataset, and Section~\ref{prompt_alg} introduces our approach to remove prompt overlap. Section ~\ref{expt_setup} presents our experimental setup, and experimental results are presented in Section~\ref{expt_lm}. Section~\ref{discussion} provides a discussion of our work, and Section~\ref{conclusion} concludes the paper.

\section{Relation to prior work}
\label{prior_work}
Dysarthric ASR using the TORGO dataset has been well studied in the literature. An ASR system in general consists of an decoder that uses an acoustic model and a language model to return a hypothesized word string. For acoustic modelling, Espana-Bonet et al.~\cite{espana2016automatic} looked at the impact of deep neural network based acoustic models on the TORGO dataset. Joy et al.~\cite{joy2018improving} look at different configurations of acoustic models to suggest improvements. Furthermore, Hermann et al.~\cite{hermann2020dysarthric} study the impact of lattice-free MMI and how it compensates for speakers with slow speaking rates. Yue et al.~\cite{yue2020exploring} state that results for language models trained on training transcripts overestimate ASR results on TORGO due to prompt overlap and investigate the use of out-of-domain language models.  In our early experiments on EC for dysarthric ASR, our models started to memorize prompts. Prompt overlap in TORGO is a serious issue we tackle first, before understanding the impact of EC.

To the best of our knowledge little or no work exists on the impact of error-correction following first-pass ASR for atypical speakers. For typical speech, sequence-to-sequence models for error-correction~\cite{dutta2022error} have shown to improve ASR performance. Furthermore, Leng at al.~\cite{leng2023softcorrect} incorporate efficient approaches to consider multiple different hypotheses for EC using transformer based encoder-decoder models. More recently cross-modal EC systems based on large-language models (LLMs) such as Whispering-LLaMA~\cite{radhakrishnan-etal-2023-whispering} have shown promise. Li et al.~\cite{li2024crossmodal} study the impact of cross-modal EC using discrete-speech units and their impact on LR-OOD tasks. In the modern deep learning literature  Park et al.~\cite{park2022pictalky} use a sequence-to-sequence EC module on Korean atypical speech data after using a commodity ASR transcription system. Similar to our work theirs is an AAC application to help children with language disabilities. In contrast to their work, our work looks at the impact of modern multi-modal generative error-correction techniques for English dysarthric ASR. 

\section{The TORGO dataset}
\label{ec_intro}
This section briefly introduces the TORGO dataset, and highlights the issue of prompt overlap. Furthermore this section talks about the leave-one-speaker out protocol for TORGO that requires speaker specific train and test sets.

The TORGO data set consists of eight speakers each with varying levels of dysarthria. The database also consists of 7 control speakers. There is about 15 hours of audio from all of the speakers, with a total of 6 hours of audio coming from dysarthric speakers, and 9 hours of audio coming from control speakers.
\begin{table}[h]
\centering
\caption{Dysarthric speakers in the TORGO dataset}
\begin{minipage}[b]{0.8\linewidth}
  \small
  \begin{tabular}{|c|c|c|c|}
    \hline
    Severity & Speaker & \# Utterances & \% Prompt Overlap \\ \hline \hline
    \multirow{4}{*}{Severe} & F01  & 228  & 100\% \\ \cline{2-4}
                            & M01  & 739  & 99.1\% \\ \cline{2-4}
                            & M02  & 772  & 98.2\% \\ \cline{2-4}
                            & M04  & 659  & 98.2\% \\ \hline
    M/S                    & M05  & 610  & 98.9\% \\ \hline
    Moderate               & F03  & 1097 & 95.7\% \\ \hline
    \multirow{2}{*}{Mild} & F04  & 675  & 98.6\% \\ \cline{2-4}
                           & M03  & 806  & 99.7\% \\ \hline    
  \end{tabular}
\end{minipage}
\label{table:torgo}
\end{table}

The database consists of recordings of single words, sentences, and descriptions of contents in photographs. The speakers, their severity levels and the number of utterances has been summarized in Table~\ref{table:torgo}. The presence of only 957 unique texts among the total 16,394 text entries indeed indicates a significant level of overlap in the prompts used by the speakers. Table~\ref{table:torgo} also summarizes this as a percentage overlap between speakers. About 75\% of the utterances are isolated word utterances. 

Our work stays consistent with the protocol defined in Espana-Bonet et al.~\cite{espana2016automatic} which has been adopted in subsequent work~\cite{hermann2020dysarthric, yue2020exploring} which uses a leave-one-speaker-out evaluation protocol. Data from speaker F03 is used as validation data, and F04 in-case the training speaker is F03. In the leave-one-speaker-out protocol the training data consists of data from all speakers other than the target speaker. This includes data from the control speakers. The data from the target speaker is used as test data. 

\section{Removing prompt overlap in TORGO}
\label{prompt_alg}
Our initial attempts involved reducing prompt overlap by manually selecting utterances. Instead of relying on heuristics, this section talks about a mixed integer linear programming (MILP) algorithm for data selection so that utterances with the same text prompt do not appear between the TORGO training and test speaker sets.  The objective of Mixed Integer Linear Programming (MILP) is to maximize or minimize a linear objective function subject to bounds, linear and integer constraints on some variables. 

Table ~\ref{variables_milp} introduces some of the variables that are useful for the MILP optimization problem. As the table mentions, binary variables $x_i$ and $y_i$ are used to filter speaker specific TORGO train and test sets $T_\text{train}$ and $T_\text{test}$. As a result of this filterting, the MILP algorithm produces $S_\text{train}$ and $S_\text{test}$, which are the speaker specific output filtered data sets without prompt overlap. We call the resulting train and test dataset, $S_\text{train}$ and $S_\text{test}$, the NP-TORGO (no-prompt overlap TORGO) dataset. 

\begin{table}[h]
\caption{A summary of variables used for mixed-integer linear programming to produce the NP-TORGO dataset}
\begin{tabular}{|c|p{6.25cm}|}
\hline
\textbf{Variable} & \textbf{Description} \\
\hline
$T_{\text{train}}$ & Original training set with prompt overlap  (all but the target speaker).\\
\hline
$T_{\text{test}}$ & Original test set with prompt overlap (only the target speaker).\\
\hline
$f$ & Fraction of test data to retain. \\
\hline
$S_{\text{train}}$ & Training set without prompt overlap, as a result of MILP selection. \\
\hline
$S_{\text{test}}$ & Test set without prompt overlap, as a result of MILP selection. \\
\hline
$x_i$ & A (binary) integer variable: 1 if prompt $i$ is in $S_{\text{train}}$, 0 otherwise. \\
\hline
$y_i$ & A (binary) integer variable: 1 if prompt $i$ is in $S_{\text{test}}$, 0 otherwise. \\
\hline
\end{tabular}
\label{variables_milp}
\end{table}
The objective function for the MILP is to maximize the number of prompts in both training and test sets:
\begin{align}
\text{Maximize} & \quad \left(\sum_{i \in T_\text{train}} x_i + \sum_{i \in T_\text{test}} y_i \right), 
\end{align}
subject to the following constraints,
\begin{enumerate}
\item No overlaps: If a prompt exists in the training set ($x_i=1$), it cannot exist in the test set ($y_i = 0$) (and vice versa).
\begin{equation}
x_i + y_i \leq 1 \quad \forall i \in (T_\text{train} \cup T_\text{test})
\end{equation}
\item Each prompt in the train and test sets must either be 1 or 0:
\begin{align}
x_i, y_i &\in \{0,1\} 
\end{align}
\item A floor constraint is set to ensure a minimum number of prompts in the test set. This ensures that the size of $S_\text{test}$ is at least a fraction $f$ of the original test set $T_\text{test}$. At $f=0$, no test prompts will be retained as the number of total prompts is maximized when no prompts are removed from the training set.
\begin{equation}
|S_\text{test}| = \sum_{i \in T_\text{test}} y_i \geq f \times |T_\text{test}|
\end{equation}
\end{enumerate}
To ensure a balanced distribution of both words and phrases in the train and test sets, the optimization problem was divided into two separate tasks: one for splitting isolated words and another for splitting phrases. The results from these two tasks were combined to construct the NP-TORGO dataset. The optimization problem is solved using the OR-Tools package developed by Google \cite{ortools}, particularly its \texttt{pywraplp} module. Additionally,  the SCIP (Solving Constraint Integer Programs) algorithm, integrated within OR-Tools was employed to effectively solve the linear programming problem. SCIP is renowned for its capability in handling mixed-integer linear programming tasks.

\begin{figure}[h]
    \centering
    \begin{tikzpicture}    
        \begin{axis}[
            xlabel={\( f \)},
            ylabel={Prompt Count},
            width=0.4\textwidth,
            height=0.3\textwidth,
            xmin=0, xmax=1,
            ymin=0, ymax=18000,
            yticklabel style={/pgf/number format/fixed}, 
            scaled y ticks=false, 
            legend pos=south west,
        ]
        
        \addplot[style=solid, color=blue, line width=0.5pt] table[x=f_value,y=M01_train,col sep=comma] {experiment_summary.csv};
        \addplot[style=solid, color=red, line width=0.5pt] table[x=f_value,y=M01_test,col sep=comma] {experiment_summary.csv};
        \addplot[style=solid, color=black, line width=1pt] table[x=f_value,y=M01_total,col sep=comma] {experiment_summary.csv};
        
        \addplot[mark=none, color=gray, style=dashed, line width=1pt] coordinates {(0.55,0) (0.55,10407)};
        \node[anchor=south west] at (axis cs:0.65,12500) {\( f=0.55\)};
        \draw[-] (axis cs:0.65,12500) -- (axis cs:0.55,10407);
        
        \legend{
            train,
            test,
            total,
        }
        \end{axis}
    \end{tikzpicture}
    \caption{Effect of \( f \) on the number of utterances retained in training and test set for target speaker M01.}
    \label{fig:promptcount}
\end{figure}
In our early manual attempts to reduce overlap for each speaker we noticed a steep drop-off in training data quantity while trying to retain enough test data. Using heuristics across different speakers we found that a large part of the training data could be retained while preserving about 60\% of the test data.  Our work chooses $f=0.55$ for the MILP algorithm so as to maintain at least 55\% of the test speaker data. Figure~\ref{fig:promptcount} shows the variation in the prompt count with $f$ for all prompts, training prompts and test prompts for one example speaker M01 in the TORGO dataset. The results of the train/test split optimization for the combined (containing both isolated words and sentences) dataset NP-TORGO are presented in Table \ref{tab:combined}. It is notable that the optimization process achieved a ratio of retained prompts in the train set, ranging from 63.3\% to 64.3\%. Similarly, the test set maintained a percentage of retained prompts between 55.0\% and 55.2\%, meeting the constraint while maximizing the available data.

\begin{table}[h]
\vspace{0.25cm}
\centering
\caption{Controlling All Prompt Overlaps}
\label{tab:combined}
\begin{tabular}{|c|c|c|c|c|c|c|c|c|c|}
\hline
{\textbf{Speaker}} & \multicolumn{2}{c|}{\textbf{Train Set}} & \multicolumn{2}{c|}{\textbf{Test Set}}
\\ \cline{2-5}
 & \textbf{Before} & \textbf{After} & \textbf{Before} & \textbf{After} \\ \hline
F01 & 16166 & 10232 & 228 & 126 \\ \hline
F03 & 15319 & 9851 & 1075 & 592 \\ \hline
F04 & 15727 & 10034 & 675 & 367 \\ \hline
M01 & 15655 & 10002 & 739 & 407 \\ \hline
M02 & 15628 & 9991 & 766 & 423 \\ \hline
M03 & 15594 & 9976 & 800 & 442 \\ \hline
M04 & 15742 & 10042 & 652 & 360 \\ \hline
M05 & 15821 & 10077 & 573 & 316 \\ \hline
\end{tabular}
\vspace{-5mm}
\end{table}

\section{Experimental setup}
\label{expt_setup}
This section talks about the acoustic models, language models and error-correction models used in the experimental study.
\subsection{Acoustic models}
Our initial experiments investigated the impact of different variations of the wav2vec2~\cite{baevski2020wav2vec} architecture. 
Preliminary results showed strong baseline performance with cross-lingual representations. The `wav2vec2-xlsr-53'~\cite{conneau2020unsupervised} model consists of cross-lingual representations learned from 53 different languages using a semi-supervised objective function. This observation is similar to results reported by Hernandez et al~\cite{hernandez2022cross}.

For our acoustic model training our setup uses the HuggingFace training tools\footnote{www.huggingface.co}. Per speaker acoustic models were trained with a connectionist temporal classification (CTC)~\cite{graves2006connectionist} objective. The symbol set consists of a dictionary of 32 tokens encompassing a range of symbols including standard alphabetic characters (`a' through `z') and a few special tokens. The acoustic models were trained using the Adam optimizer on an NVIDIA T4 GPU with a batch size of 4 and gradient accumulation every 2 steps. The learning rate was set to $10^{-4}$ with a linear warmup of 1000 steps. Regularization was applied using weight decay of $5 \times 10^{-3}$. While training was set to 20 epochs, the best model was chosen based on loss and word error rate scores on the validation set.

Additionally, to understand off-the-shelf naive model inference we use a `wav2vec2-xlsr-53-en'\footnote{https://huggingface.co/jonatasgrosman/wav2vec2-large-xlsr-53-english} model trained on the Common-Voice dataset and we contrast this with the performance of naive inference with the Whisper-Large-V2 model.

\subsection{Language Models}
For training n-gram language models (LMs) our work uses the KenLM\cite{heafield2011kenlm} toolkit. By default, KenLM utilizes modified Kneser-Ney smoothing including interpolation with weight backoff. All of our LM evaluations use different \textit{tri-gram} language models. This includes an in-domain language model called \textbf{TORGO LM}, and out-of-domain language models called \textbf{NP-TORGO LM} and \textbf{Librispeech LM}.

For the \textbf{TORGO LM} and \textbf{NP-TORGO LM}, a different tri-gram LM is trained for each evaluation speaker. The training text consists of prompts from all other speakers in the training set except the test speaker from the TORGO and the NP-TORGO datasets respectively. For example, while choosing F01 as test speaker, F03 as default validation speaker, the training data consists of texts from all speakers except F01 and F03. 

For the tri-gram \textbf{Librispeech LM}, text from the 360h training subset of the LibriSpeech corpus~\cite{panayotov2015librispeech} is used. Librispeech is a read speech dataset based on LibriVox’s audio books. There are fifty-eight thousand unique words, one-hundred thousand sentences, and 3.6 million tokens to train the language model.

Our analysis uses both the out-of-vocabulary (OOV) rate as well as perplexity. Table~\ref{tab:combinedStats} gives the OOV rate and perplexity for the test set for each speaker in the NP-TORGO dataset. In contrast Table ~\ref{tab:in_domain} gives the perplexity and OOV rate for the original TORGO dataset. It is apparent that removing prompt overlap yields a high OOV rate for the NP-TORGO dataset.
\begin{table}[h]
\centering
\caption{Out-of-domain LMs : Avg. Perplexity and OOV rate}
\label{tab:combinedStats}
\begin{tabular}{|c|c|c|}
\hline
\textbf{Trained LM} & \textbf{Perplexity} & \textbf{OOV rate} \\ \hline
LibriSpeech & 3979.84 & 2.47\% \\ \hline
NP-TORGO & 462.97 & 59.99\% \\ \hline
\end{tabular}
\vspace{2mm}
\caption{In-domain LM : Avg. Perplexity and OOV rate}
\label{tab:in_domain}
\begin{tabular}{|c|c|c|}
\hline
\textbf{Trained LM} & \textbf{Perplexity} & \textbf{OOV rate} \\ \hline
TORGO & 19.24 & 0.63\% \\ \hline
\end{tabular}
\vspace{-5mm}
\end{table}

\subsection{Cross-modal error-correction}
In order to study the impact of acoustic input for post-processing ASR, our study uses a recently proposed cross-modal error-correction (EC) model called Whispering-LLaMA~\cite{radhakrishnan-etal-2023-whispering}\footnote{Code taken from https://github.com/Srijith-rkr/Whispering-LLaMA}. As the name suggests, this EC model consists of the acoustic encoder from Whisper~\cite{radford2023robust} with cross-modal attention to adapters in the LLaMA~\cite{touvron2023llama} model. 

The authors introduce per layer adapter modules to the frozen LLaMA model. The first adapter variable $A_{i}^{L}$ represents the adapter used in layer $i$ to fine-tune the LLaMA model using a scaled dot product attention mechanism. The second adapter variable $A_{i}^{W}$ refers to another adapter in layer $i$ used to fuse Whisper features with the LLaMA model following an autoencoder mechanism. Each of these adapters have learnable matrices $\textcolor{red}{M_{\theta}^{i}}$ for the adapter variable $A_{i}^{L}$ and $\textcolor{red}{M_{down}^{i}}$ and $\textcolor{red}{M_{up}^{i}}$ for the cross-modal fusion adapter variable $A_{i}^{W}$. The authors provide a parameter $r$ which can be chosen as 8,16 or 32 to adjust the sizes of $\textcolor{red}{M_{down}^{i}}$ and $\textcolor{red}{M_{up}^{i}}$. These respectively yield small, medium or large adapters. In our experiments we tried small(r=8) and medium (r=16) adapters. 

The adapters are fine-tuned using instruction-fine tuning with n-best hypotheses generated by a Whisper model. Our experiments do not fine tune the Whisper model at all. In the original paper the authors use the Whisper-tiny model to generate weak hypotheses for the n-best list. For dysarthric speech it was found that we needed hypotheses generated by a larger Whisper-Large-V2 model. The number of hypotheses set to $n=50$, instead of $n=200$ used in the paper in order to save computation time. The acoustic features for cross-modal fusion were generated with the Whisper-Large-V2 model as well.

The LLaMA model in this implementation has a maximum sequence length of 2048 tokens, and a maximum input length of 1000 tokens. We trained the adapters for 10 epochs, with a weight decay of 0.02, warmup set to 0, batch-size of 32 and a micro batch-size of 4 with a linear learning rate decay strategy.

\section{Experiments}
\label{expt_lm}
This section summarizes the results of our experimental study. Table~\ref{tab:word_error_rate} summarizes our baseline results with the wav2vec2-xlsr-53 model trained on dysarthric data with greedy CTC decoding. The results are presented on the TORGO dataset and the NP-TORGO dataset on a per speaker basis. For the TORGO dataset, results are presented on the full test set (100\% test) and a reduced test set (55\% test). The reduced test set (55\% test) is identical to the test set obtained when constructing the NP-TORGO dataset. \begin{table}[h]
\centering
\caption{Baseline wav2vec2-xlsr53 with greedy decoding}
\label{tab:word_error_rate}
\resizebox{\columnwidth}{!}{
\begin{tabular}{|c|c|c|c|c|}
\hline
{\textbf{Severity}} & \textbf{Speaker} & \multicolumn{2}{c|}{\textbf{TORGO}} & {\textbf{NP-TORGO}}
\\ \cline{3-4}
 & & \textbf{100\% Test} & \textbf{55\% Test} & \\ \hline
\text{Severe}   & F01 & 46.45\% & 47.27\% & 80.00\% \\ \cline{2-5} 
                & M01 & 40.72\% & 43.28\% & 85.07\% \\ \cline{2-5} 
                & M02 & 54.40\% & 55.21\% & 90.43\% \\ \cline{2-5} 
                & M04 & 64.50\% & 65.81\% & 93.75\% \\ \hline
\text{M/S}      & M05 & 51.04\% & 56.20\% & 90.38\% \\ \hline
\text{Moderate} & F03 & 25.81\% & 25.43\% & 65.14\% \\ \hline
\text{Mild}     & F04 &  4.92\% &  4.46\% & 45.06\% \\ \cline{2-5} 
                & M03 &  3.17\% &  3.76\% & 45.21\% \\ \hline
\end{tabular}}
\vspace{-3mm}
\end{table}

Tables~\ref{tab:torgo_lm} and ~\ref{tab:np_torgo_lm} break down the performance by severity on the TORGO and the NP-TORGO datasets respectively. Furthermore, the impact of various language models is presented at the isolated word (IW) level as well as the sentence (Sent.) level. Each row describes the impact on the word error rate (WER) of a particular language model. Table~\ref{tab:np_torgo_lm} additionally contains results for naive inference with off-the-shelf wav2vec2-xlsr-53 (labeled as w2v2-xlsr53-en) models as well as the Whisper-Large-V2 (labelled as Whisper-L-V2) model. Furthermore, the last two rows of Table~\ref{tab:np_torgo_lm} list the performance of the Whispering-LLaMA EC models with small (labelled as WL EC (s)) and medium adapters (labelled as WL EC (m)).
\begin{table}[h]
  \centering
  \caption{Isolated word (IW) and Sentence (Sent.) performance using various language models on TORGO and NP-TORGO}
  \label{tab:torgo_lm}
  \resizebox{\columnwidth}{!}{%
    \begin{tabular}{lcccccc}
      \toprule
      \multicolumn{7}{c}{\textbf{TORGO Dataset}}\\
      \toprule
      \textbf{Task} & \multicolumn{2}{c}{\textbf{Severe}} & \multicolumn{2}{c}{\textbf{Moderate}} & \multicolumn{2}{c}{\textbf{Mild}} \\
      \cmidrule(lr){2-3} \cmidrule(lr){4-5} \cmidrule(lr){6-7}
      & \textbf{IW} & \textbf{Sent} & \textbf{IW} & \textbf{Sent} &  \textbf{IW} & \textbf{Sent} \\
      \midrule
      Baseline & 54.9\% & 50.0\% & 51.4\% & 30.8\% & 7.0\% & 2.5\% \\
      TORGO LM & 49.05\% & 37.9\% & 45.4\% & 23.4\% & 6.7\% & 1.6\% \\
      Librispeech LM & 51.3\% & 42.9\% & 47.9\% & 24.5\% & 6.8\% & 1.8\% \\
      \bottomrule
    \end{tabular}%
  } 
\end{table}

\begin{table}[h]
  \vspace{0.25cm}
  \centering
  \caption{Performance on NP-TORGO}
  \label{tab:np_torgo_lm}
  \resizebox{\columnwidth}{!}{%
    \begin{tabular}{lcccccc}
      \toprule
      \multicolumn{7}{c}{\textbf{NP-TORGO Dataset}}\\
      \toprule
      \textbf{Task} & \multicolumn{2}{c}{\textbf{Severe}} & \multicolumn{2}{c}{\textbf{Moderate}} & \multicolumn{2}{c}{\textbf{Mild}} \\
      & \textbf{IW} & \textbf{Sent} & \textbf{IW} & \textbf{Sent} &  \textbf{IW} & \textbf{Sent} \\
      \midrule
      w2v2-xlsr53-en & 114.8\% & 73.9\% & 95.9\% & 40.2\% & 41.7\% & 8.42\%\\
      Whisper-L-V2 & 108.1\% & 56.3\% &88.4\%& 21.8\% & 28.9\% & 5.16\%\\
      Baseline & 93.9\% & 83.8\% & 93.9\% & 68.6\% & 68.4\% & 35.6\% \\
      NP-TORGO LM & 96.2\% & 80.0\% & 95.2\% & 61.7\% & 73.6\% & 33.8\% \\
      Librispeech LM & 93.9\% & 78.5\% & 90.0\% & 58.8\% & 63.5\% & 27.8\% \\
      WL EC (s) & 92.9\%& 79.1\%& 86.1\% & 28.8\% & 55.1\% & 13.8\%\\
      WL EC (m) & 93.0\% & 62.9\% & 84.6\% & 29.1\% &50.0\% & 11.3\%\\
      \bottomrule
    \end{tabular}%
  }
  \vspace{-5mm}
\end{table}
\vspace{-0.25cm}

\section{Discussion}
\label{discussion}
From the results in Table~\ref{tab:word_error_rate} it appears that cross-lingual representations using the wav2vec2-xlsr-53 model provide a new state-of-the-art performance on TORGO (100\% Test). The WER performance does not appear to be drastically affected when presented on the reduced test set (55\% test). The improvements are especially apparent when looking at the mild speakers, with their WER performance in the 3-5\% range. 

Prompt overlap leads to a significant over-estimation of the performance, where the model easily starts to memorize all of the prompts. This is consistent with the observations made by Yue et al.~\cite{yue2020exploring}. On removing prompt overlap, the same acoustic models perform rather poorly overall on the NP-TORGO dataset. The impact is evident when it comes to isolated word recognition. For speaker M05 for example, the words recognized are acoustically close to the ground truth, but are not the exact word. With the drastic drop in performance for NP-TORGO,  there is strong evidence to suggest that prompt overlap leads to an over-estimation of the WER performance on TORGO. 

A further breakdown of the results on the TORGO and NP-TORGO datasets appears in Tables~\ref{tab:torgo_lm} and ~\ref{tab:np_torgo_lm}. In Table~\ref{tab:torgo_lm} in-domain tri-gram LMs with backoff are seen to have a significant impact on isolated word WER performance for speakers with severe and moderate dysarthria. The impact of out of domain language models (such as Librispeech LM) appears limited. On the other hand in NP-TORGO performance in Table~\ref{tab:np_torgo_lm} suggests that out-of-domain language models have a more significant role to play when there is no prompt-overlap between train and test utterances. Out-of-domain language models are seen to have a limited impact on the isolated word performance, and understandably a larger impact on the sentence level performance for severe, moderate and mild speakers. In addition the prompts taken from the NP-TORGO training set to train the NP-TORGO LM seem to hurt the performance on the isolated word recognition task.

Naive inference results in Table~\ref{tab:np_torgo_lm} for off-the-shelf models such as Whisper-Large-V2 and the Wav2vec2-xlsr-53-en model show poor performance on the isolated word task, but provide some strong results for moderate and naive speakers on the sentence level task. Performance on mildly dysarthric speakers seems to be outstanding, but less so for those with severe dysarthria. Results with our baseline model where dysarthric speaker data without prompt overlap is used to train our (`wav2vec2-xlsr-53`) acoustic model seem to yield moderate performance improvements in the case of speakers with severe dysarthria on the isolated word task. Error-correction using the Whispering-LLaMa models show some promise. They are able to improve on the isolated word task performance for severe speakers, but degrade sentence level performance compared to naive inference.  The performance for mild speakers seems to get worse for sentence level correction compared to naive inference. 
\section{Conclusion}
\label{conclusion}
Our work shows that state-of-the-art models even after post-processing with powerful error-correction models are still not ready for servicing dysarthric speakers in telehealth and in-person healthcare settings. To evaluate existing models this work presented a principled method for constructing a dysarthric ASR dataset as a subset of TORGO without prompt-overlap using linear programming. Results were presented to understand the impact of performing this split. Furthermore, our results show that cross-modal error-correction models based on LLMs hold promise for dysarthric ASR, but they struggle when it comes to isolated word scenarios. In order to improve isolated and sentence level word performance on this challenging baseline, future work will look at incorporating phonetic information into error-correction models along with careful data augmentation. 

\section*{Acknowledgment}

The authors would like to acknowledge support from the Khoury West Coast Research Fund and the Khoury SEED Grant from the Khoury College of Computer Sciences from Northeastern University that financially supported Macarious Hui and Jinda Zhang for this work.



\bibliographystyle{IEEEtran}
\bibliography{IEEEabrv,refs}
%

\end{document}